\documentclass[11pt,a4paper]{article}

\usepackage[top=12mm,bottom=12mm,left=30mm,right=30mm,head=12mm,includeheadfoot]{geometry}
\usepackage[utf8]{inputenc}
\usepackage{amsmath}
\usepackage{graphicx}
\usepackage{xcolor}
\usepackage{cite}
\usepackage[width=.90\textwidth]{caption}
\usepackage[nottoc,notlot,notlof]{tocbibind}
\usepackage{fancyhdr}
\usepackage{lineno}
\usepackage{hyperref}
\usepackage{doi}

\definecolor{titleblue}{HTML}{002B49}

\hypersetup{
    colorlinks,
    linkcolor={red!50!black},
    citecolor={blue!50!black},
    urlcolor={blue!80!black}
}

\IfFileExists{mathdesign.sty}{%
  \usepackage[bitstream-charter]{mathdesign}%
}{%
  \usepackage{charter}%
  \usepackage{amssymb}%
}
\usepackage{booktabs}
\usepackage{makecell} 
\usepackage{xspace}
\usepackage{subcaption}
\usepackage[most]{tcolorbox}

\newtcolorbox{promptbox}[1]{%
  enhanced, breakable,
  colback=gray!5, colframe=gray!50, boxrule=0.5pt, arc=2pt,
  left=6pt, right=6pt, top=4pt, bottom=4pt,
  fonttitle=\small\sffamily\bfseries, coltitle=black, colbacktitle=gray!15,
  title={#1},
  fontupper=\small\ttfamily, before upper={\raggedright\setlength{\parskip}{0.5\baselineskip}},
}

\DeclareSymbolFont{usualmathcal}{OMS}{cmsy}{m}{n}
\DeclareSymbolFontAlphabet{\mathcal}{usualmathcal}

\fancypagestyle{plain}{
\fancyhf{}

\fancyfoot[C]{\textbf{\thepage}}
}

\title{Goal-Persistent Coding Agents as Scientific Performance Engineers:\\
A Fixed-Radius Nearest-Neighbor Case Study}
\author{Draft for the \texttt{libFRNN} project}
\date{September 2026}

\begin{document}
\pagestyle{plain}
\begin{center}{\Large \textbf{\color{titleblue}{
Goal-Persistent Coding Agents as Scientific Performance Engineers:\\
A Fixed-Radius Nearest-Neighbor Case Study
}}}\end{center}

\begin{center}\textbf{
Xiangyang Ju\textsuperscript{1$\dagger$}
}
\end{center}

\begin{center}
{\bf 1} Lawrence Berkeley National Laboratory, Berkeley, CA, USA\\[3pt]

$\dagger$ \href{mailto:xju@lbl.gov}{\small xju@lbl.gov}
\end{center}

\section*{\color{titleblue}{Abstract}}
\boldmath\textbf{
Coding agents can pursue persistent objectives across many tool-use turns, but
evidence that general-purpose agents can conduct rigorous scientific
performance engineering remains limited.  We present a repository-scale case
study in which off-the-shelf Codex and Claude Code agents optimize
fixed-radius nearest-neighbor (FRNN) search for particle tracking.  Starting
from a PyTorch-dependent CUDA implementation, the agents follow an executable
goal that specifies exact-correctness tests, profiling requirements, and
acceptance criteria without prescribing code transformations.  In the primary
sequential trajectory, they autonomously remove the PyTorch dependency and conduct
hypothesis-driven optimization experiments.  The resulting
standalone C++/CUDA library exactly reproduces the targeted reference result.
Its synchronous NumPy interface achieved 1.6-fold speedup over the original GPU-resident
PyTorch interface, despite including host transfers.
Similar speedups were observed across different GPU architectures and software stacks.
An independent optimization rerun followed a different sequence of
hypotheses and reached even better performance on the target workload.
These results show that goal-persistent coding agents can act
as experimental performance engineers, and that executable scientific
contracts are needed both to guide and to validate their optimization.
}

\vspace{\baselineskip}


\section{Introduction}
\label{sec:intro}

Optimizing scientific GPU software is often challenging because it requires
both domain knowledge and a detailed understanding of the software and hardware stack.
A senior performance engineer uses profiling tools to
identify computational bottlenecks, formulates hypotheses about their causes, and
implements changes to test those hypothesis.
Based on the measured results, the engineer retains, revises,
or rejects each change.
This process is inherently iterative:
later experiments are informed by the outcomes of earlier ones,
and many hypothesis-driven cycles may be required before a satisfactory solution is reached.

Recent goal-persistent coding agents, such as Codex~\cite{codexgoals2026} and Claude Code~\cite{claudegoals2026},
can maintain an explicit completion contract across many tool-use turns, while
repository artifacts provide durable
memory across context windows~\cite{anthropicharness2025}.  Whether a
general-purpose coding agent can use these mechanisms to carry out credible scientific
performance engineering remains an open empirical question.

We study that question through fixed-radius nearest-neighbor (FRNN) search.  For
query points $x_i\in\mathbb{R}^D$ and reference points $x_j$, FRNN returns
\begin{equation}
  \mathcal{N}_r(i)=\{j:\lVert x_i-x_j\rVert_2\leq r\},
\end{equation}
optionally retaining at most $K$ neighbors in distance order.  Unlike
$k$-nearest-neighbor search, FRNN preserves a physical or learned length scale
and permits the degree to vary with local density.  FRNN is used in many
scientific applications, including
interaction lists in molecular and colloidal simulation
\cite{howard2016molecular}, local kernels in smoothed particle hydrodynamics
\cite{winkler2018sph}, and ball queries in learned point-cloud models
\cite{qi2017pointnetpp}.  It is also a graph-construction bottleneck in the
ExaTrkX particle-tracking pipeline~\cite{ju2021exatrkx} considered here.

FRNN provides a useful test case for agentic optimization because correctness and performance
pull in different directions.  Brute force costs $O(N^2D)$ but provides exact results.
Uniform grids, cell lists, trees, and spatial sorting reduce the
candidate set and often achieve near-linear empirical scaling at bounded
occupancy, yet their behavior depends strongly on dimension, density, radius,
and the maximum neighbor count. Moreover, seemly minor changes to floating-point accumulation
or boundary screening can reduce latency while silently changing neighbor membership
near the radius threshold.  A successful optimization agent must therefore optimize hard-level
performance while preserving an exact scientific correctness contract.

We ask whether off-the-shelf, general-purpose coding agents can sustain the
complete optimization loop over a real repository, beyond emitting a fast
kernel in a single response.  The agents
receive a persistent goal containing correctness constraints, evaluation
procedures, and candidate optimization areas, but no prescribed transformation
sequence.  They choose successive hypotheses, execute experiments on hardware,
and use the observed evidence to determine what to try next, in autonomous loops.

This paper makes three contributions:
\begin{enumerate}
  \item a repository-scale case study of goal-persistent, hypothesis-driven
        optimization with general-purpose Codex and Claude Code agents.
  \item an analysis of the executable goal, repository artifacts, profiler
        feedback, and handoff state that support a multi-session optimization
        trajectory without a purpose-built autoresearch controller; and
  \item an exact standalone C++/CUDA FRNN implementation, evaluated on
        controlled inputs and a realistic high-dimensional particle-tracking
        embedding.
\end{enumerate}
The primary trajectory establishes feasibility, while the continuation-stage
rerun illustrates an alternative optimization path and clarifies the workload
assumptions encoded by the executable contract.  Controlled comparisons across
repeated runs and alternative optimization approaches remain future work.

\section{Related work}

Fixed-radius nearest neighbor search is a workhorse subroutine across
computational science, from molecular model-building~\cite{levinthal1966molecular}
to cosmological catalogs~\cite{chalela2021grispy}.  It also builds the
candidate graph in the ExaTrkX geometric deep-learning pipeline for
charged-particle tracking~\cite{ju2021exatrkx}, where a dedicated acceleration
study~\cite{lazar2022exatrkxacceleration} cut per-event inference from roughly
$15$ s to $0.7$ s by combining cuGraph, mixed precision, and Xue's FRNN for
graph construction, putting the kernel studied here on the critical path of a
production reconstruction chain.  The real-world example used below, with
271,663 points in 12 dimensions, is representative of that ExaTrkX-style
embedded-space graph construction.

Our study starts from Xue's FRNN
implementation~\cite{xue2020frnn,xue2020slides}, referred to as
the original FRNN in this paper,
which uses a uniform-grid CUDA kernel in the
Hoetzlein lineage exposed through a \texttt{pytorch3d}-compatible interface,
reporting $12$--$30\times$ speedups over \texttt{pytorch3d}'s $k$-nearest-neighbor
operator and recommending brute force below roughly $10^4$ points.  Its design
notes observe that fixed-radius search has lower thread divergence than $k$-NN
and is more robust to variation in point density, and that the grid spacing
should be chosen so that the search radius is an integer multiple of it.
\texttt{libFRNN}~\cite{libfrnn2026} rewrote Xue's implementation in C++,
keeping the PyTorch dependency, the host-side synchronization, and the
array-of-structures memory layout.

\subsection{Agentic code optimization}
\label{sec:agent-related}

LLM-based optimization systems fall into two overlapping families.  The first
uses explicit population-based search,
such as ReEvo~\cite{reevo2024}, AlphaEvolve~\cite{alphaevolve2025}, AdaEvolve~\cite{adaevolve2026}, and AVO~\cite{avo2026}.
They maintain evolutionary
databases of programs, sample prior high-scoring candidates in prompts, and
use executable evaluators to select new variants across mathematical,
algorithmic, and systems problems.  These systems gain parallel diversity and a
durable archive, but they need a purpose-designed search and evaluation
harness.

The second family more directly imitates a performance engineer.  CudaForge
separates a code-producing agent from a judge that interprets correctness and
selected Nsight Compute metrics \cite{cudaforge2025}; KForge similarly combines
a synthesis agent with a performance-analysis agent and iterates through
functional and optimization passes on CUDA and Metal \cite{kforge2025}.
FormulaCode extends evaluation to real repository-scale performance tasks and
shows that frontier agents obtain nontrivial speedups but still underperform
expert patches under multi-workload evaluation \cite{formulacode2026}.
Autonomous-science systems cover the wider research loop.  AI Scientist-v2 formulates hypotheses,
executes experiments, analyzes results, and uses agentic tree search
\cite{aiscientistv2}, while ScienceFlow emphasizes recoverable executable
states and budget-aware control over long research horizons
\cite{scienceflow2026}.

\subsection{Autoresearch Frameworks}

Karpathy's \texttt{autoresearch} repository presents a closely related
minimal pattern: an agent repeatedly modifies training code, runs a
fixed-duration experiment, evaluates the result, keeps or discards the change,
and records the outcome while operating unattended overnight
\cite{karpathyautoresearch2026}.  It targets language-model training instead
of scientific CUDA software, but describes the same autonomous empirical loop.

Concurrent work by Takahara et al.~\cite{synagent2026} introduced SynAgent, a multimodal
multi-agent framework for hypothesis-driven autonomous materials synthesis.
SynAgent and our study share a hypothesis-driven, evidence-grounded loop but instantiate
it in complementary environments. SynAgent acts through a physical laboratory
and evolves scientific understanding, whereas our agents act on a scientific
software repository and optimize performance subject to executable correctness
constraints.

\texttt{autoresearch}, AI Scientist-v2, ScienceFlow, and SynAgent are examples
of purpose-built autoresearch frameworks.  Their harnesses encode
research-specific control structures such as fixed experiment cycles, tree
search, persistent scientific state, specialized agent roles, or
verify--falsify scheduling.  This structure can make the agents more reliable
and their research easier to follow.  However, it also makes it hard to
separate the contribution of the harness from the capability of the underlying
agent.

\section{Autonomous optimization methodology}
\label{sec:agent-method}

The primary campaign pursued two sequential
goals: first remove the PyTorch and Python dependencies, then reduce FRNN
latency while preserving brute-force-equivalent results.  We analyze a later
continuation-stage Claude rerun from the Codex handoff as a separate,
independent trajectory that complements the primary campaign.
Codex and Claude Code executed these
goals through persistent-objective mechanisms
that keep a completion contract active across many tool-use turns and require
the agent to audit completion against repository evidence.

\begin{figure}[htbp]
\centering
\includegraphics[width=\linewidth]{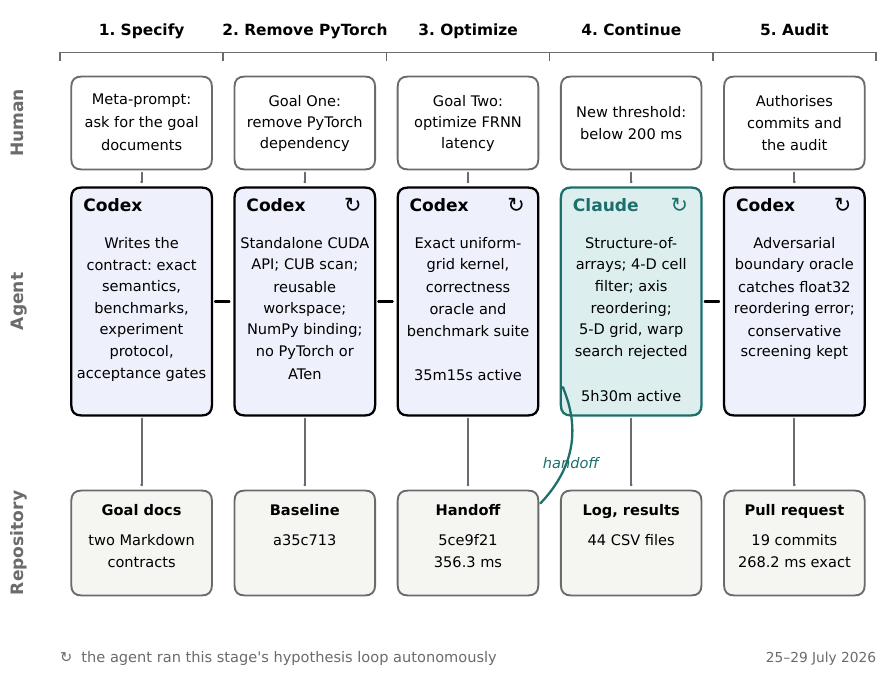}
\caption{Overview of the primary campaign.  Five stages proceed from left to
right, with the human, agent, and repository roles shown from top to bottom.
Each execution stage expands into the hypothesis loop in
Figure~\ref{fig:agent-loop}; goal construction does not.}
\label{fig:campaign}
\end{figure}

Figure~\ref{fig:campaign} summarizes the primary campaign as five stages and
three roles.  The human made only minor interventions: the meta-prompt that
asks for the goal documents, the launch of each goal, the latency threshold
that defines when the continuation goal is complete, and the authorization of
commits and of the final audit.  The agents did everything in between.  Codex
wrote the operational contract and executed the two goals; Claude Code
then continued the optimization; a later Codex audit repaired an exactness
defect that the original test suite did not expose.  The repository connects
the two agents.  Because no conversational state crossed the handoff,
we treat the primary campaign as a single trajectory.
Each marked stage expands internally into the hypothesis loop of
Figure~\ref{fig:agent-loop} without human intervention.

\subsection{From a human meta-prompt to machine-executable goals}
\label{sec:prompt-analysis}

We did not write the optimization specification ourselves.  The free-form
human instruction in the campaign was a meta-prompt asking Codex to draft it.
\begin{promptbox}{User prompt to Codex, gpt-5.6-sol, high}
\footnotesize
I'd like improve this repository
\url{https://github.com/xju2/libFRNN/tree/main} in two terms:\\
1) Remove the dependance of PyTorch\\
2) Optimize the FRNN algorithm: low latency and the same results as brutal force.\\
Now, please help me to draft a detailed prompt
so that I can utilize the "goal" capability from codex APP to achieve the two goals.
I'd like codex to work on the two goals separately, one after another.
\end{promptbox}
Codex produced two Markdown goal documents and committed them to the
repository: \texttt{docs/remove\_torch\_dep.md} and
\texttt{docs/optimize\_compute.md}.

In \texttt{docs/optimize\_compute.md}, Codex defines the starting
repository state, required outcomes, and a correctness contract.  The
goal organizes execution into three phases:
\begin{enumerate}
  \item Phase 1: Establish an immutable baseline
  \item Phase 2: Profile before redesigning
  \item Phase 3: Perform hypothesis-driven experiments
\end{enumerate}
It also identifies structural optimization families to investigate and
specifies correctness tests, performance acceptance rules, and artifacts that
the agents must preserve.

The excerpt instructing the continuation rule highlights
the demand of continuous auto-optimization.
\begin{promptbox}{Excerpt from \texttt{docs/optimize\_compute.md}}
\footnotesize
Do not stop after changing block size, compiler flags, grid resolution, or a few constants.
Those experiments are allowed,
but the work must also investigate structural bottlenecks and substantial algorithmic improvements.

Continue profiling, forming hypotheses, implementing experiments, testing correctness, and benchmarking until:
\begin{itemize}
\item the correctness contract is fully satisfied;
\item meaningful latency improvement is demonstrated;
\item the major remaining bottlenecks are understood;
\item further changes have no credible expected benefit or require an explicitly out-of-scope redesign.
\end{itemize}
\end{promptbox}

\subsection{Autonomous optimization loop}
\begin{figure}[htbp]
\centering
\includegraphics[width=0.85\linewidth]{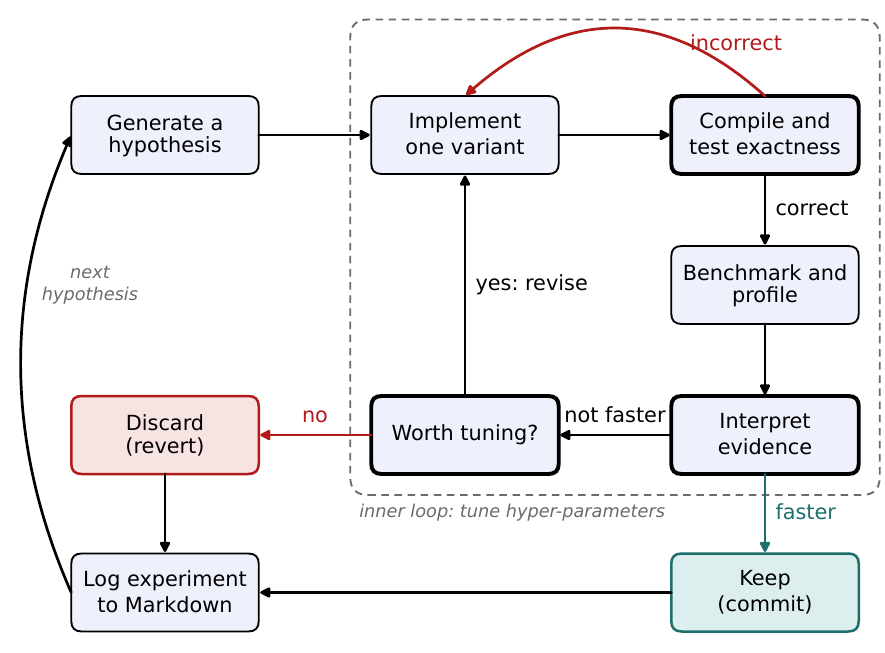}
\caption{The two-level autonomous optimization loop.  The outer loop
iterates over hypotheses.  Within the dashed region, the inner loop adjusts
implementation hyperparameters until a variant improves latency or the agent
concludes that further tuning is unlikely to help.}
\label{fig:agent-loop}
\end{figure}

The goal specification defines two nested loops (Figure~\ref{fig:agent-loop}).
The outer loop iterates over optimization hypotheses, which are meant to cover
the relevant algorithmic design choices and CUDA implementation details.

The inner loop iterates over
the hyperparameters of a single hypothesis: when a correct variant is not
faster than the incumbent, the agent decides whether the hypothesis still has
settings worth tuning, such as CUDA block size, compiler flags, or grid
resolution, and if so revises and re-evaluates it.  A variant that reduces
latency is committed and becomes the incumbent for the next hypothesis; a
hypothesis whose tunable settings are exhausted without a speedup is discarded.
Every experiment, kept or discarded, is logged.

\subsection{Executing and handing off the goals}
Each goal was launched with a single command.
Codex achieved the first goal in 18~min 40~s, which produced the standalone C++/CUDA API on which the
optimization campaign then operated.
For the second goal, Codex ran for 35~min 15~s of active time and stopped when the workspace spend
cap was reached, before the agent judged the goal complete.
By that point Codex had committed an exact implementation, the benchmark matrix, a
correctness oracle, an optimization log, and a performance report at commit
\texttt{5ce9f21}, which measured 356.3~ms on the real workload.
Finally, Codex stated explicitly that it was not yet
declaring completion, and produced no final response.

\begin{promptbox}{Prompt to Codex for Goal One, gpt-5.6-sol, high}
\footnotesize
/goal Read remove\_torch\_dep.md and execute the goal there.
\end{promptbox}

\begin{promptbox}{Prompt to Codex for Goal Two, gpt-5.6-sol, high}
\footnotesize
/goal Optimize the standalone libFRNN implementation for low latency while producing the same results
as an independent brute-force fixed-radius nearest-neighbor search. Details in optimize\_compute.md.
\end{promptbox}

We continued the same goal with Claude Code and pointed it at the
artifacts produced by Codex.
\begin{promptbox}{User prompt to Claude Code, claude-sonnet-4-6, high}
\footnotesize
Please read the docs/optimize\_compute.md for the "goal" and
docs/performance\_report.md for the current performance report from codex.
Optimization logs can be found at: docs/optimization\_log.md. Your task is to
continue the optimization by exploring different algorithmic designs and
cuda-specific improvements. You should not stop until the inference time for
the real life example using the available GPU is below 200 ms.
\end{promptbox}

This prompt adds one new piece of information, the 200~ms threshold.
We picked this number arbitrarily.  It is aggressive but not obviously
impossible for the available GPU, and we chose it because it might be
unreachable, so that the agent would keep forming hypotheses instead of
stopping at the first comfortable improvement.  The decision to mix providers
was also opportunistic.

To probe run-to-run variation, we later launched an independent Claude Code
run from the exact Codex handoff.  The run had a nominal
external wall-clock budget of 5.5~h, but the Claude Pro session-usage limit
terminated it after 91.9~min.  We report this trajectory separately.

\section{Results}

\subsection{Experimental configuration}
\label{sec:experimental-config}

The primary optimization and profiling measurements were performed on one NVIDIA
GeForce RTX 2070 SUPER (compute capability 7.5, 40 SMs, 8~GiB), driver
580.173.02, CUDA Toolkit/runtime 12.0, and CUDA compiler 13.0.88.
The follow-up Claude Code rerun uses CUDA 13.0 on the same GPU.
The cross-hardware evaluation additionally uses
NVIDIA A100-SXM4 GPUs with 40~GB and 80~GB of memory at NERSC.

The representative event has 271,663 space points with $D=12$ dimensions.
The FRNN query uses a radius of $r=0.12$ and $K=1000$ nearest neighbors.
That results in 9,279,672 directed non-self reference edges.
Original FRNN starts with device-resident PyTorch input and
returns dense device tensors. The
\texttt{libFRNN} uses a reused workspace and device-resident
input/output.
The workspace is a resuable, pre-allocated block of GPU device memory
that holds all intermediate buffers needed for FRNN search so that
no additional memory allocation is needed during search.
However, the NumPy binding in \texttt{libFRNN}
includes data transfers and compact host
output, introducing additional overhead that is not present in the original FRNN.

\subsection{Primary agent-directed optimization trajectory}

For the latency comparison between the original FRNN and \texttt{libFRNN}, shown in Table~\ref{tab:experiments},
each received three warm-up calls followed by 30 paired rounds with randomized execution order.
CUDA was synchronized immediately before and after every timed call,
and both outputs were checked against the complete
reference edge set before timing.
We report 95\% percentile confidence intervals for each
median using 10,000 resamples.  An exact two-sided sign test evaluates the
direction of the paired differences.
The variant names identify what each candidate changed in the kernel and are
kept only so the implementation can be traced back to a specific commit; the
optimization trajectory itself is carried by the latency and outcome columns,
not by the naming scheme.

\begin{table}[htbp]
\centering
\caption{Representative real-workload optimization experiments.  Each row is
a warm-device $p_{50}$.  Codex top-$K$ prototypes used 3 warm-up and 10
measured calls; all other rows used 10 warm-up and 30 measured calls.}
\label{tab:experiments}
\begin{tabular}{lrl}
\toprule
Variant & Device $p_{50}$ (ms) & Outcome \\
\midrule
Codex insertion top-$K$, SM52 & 370.772 & Pre-selection baseline \\
Codex always-heap top-$K$, SM52 & 346.110 & Rejected: sparse regression \\
Codex hybrid top-$K$ (32), SM52 & 358.924 & Superseded \\
Codex hybrid top-$K$ (24), SM52 & 357.715 & Retained design \\
Codex handoff, SM52 & 356.277 & Exact checkpoint \\
Claude baseline, SM75 & 360.341 & Session baseline\\
Structure-of-arrays layout & 340.628 & Retained \\
Four-dimensional cell filter & 315.708 & Retained \\
High-dimensional axis reordering & 299.307 & Retained as screening \\
Grouped loads and launch bounds & 260.313 & Failed later exactness audit \\
Five-dimensional grid & 408.558 & Rejected \\
Warp-cooperative search & 488.900 & Rejected \\
Codex audit: specialized screen, SM75 & 263.597 & Superseded \\
Codex audit: roundoff-safe grid, SM75 & 267.829 & Exact retained result \\
\bottomrule
\end{tabular}
\end{table}

The initial Codex phase improved the geometric-mean warm latency of five
synthetic acceptance cases by 1.80-fold and reduced conservative temporary
edge capacity on the real workload from about 4.35~GB to 148~MB through
count-first exact allocation.  The later high-dimensional work targeted the
remaining neighbor-search bottleneck.  Both trajectories are non-monotonic:
latency went up as well as down from one experiment to the next.  They cannot
be used to estimate the causal effect of goal persistence, because variants
were attempted at different repository states and the independent rerun
exhausted a provider session allowance long before its external wall-clock
budget.  The two trajectories show different hypothesis sequences and failure
modes, but they are not a matched-budget replication.

Claude's fastest candidate is 27.8\% faster than its local baseline, but it
is not a valid endpoint because reordered accumulation changed float32
membership near the radius boundary.  Conservative screening and
canonical-order recomputation retained most of the gain: the final audited
267.829~ms result is 25.7\% faster than the Claude baseline (1.35-fold
speedup).
Profiling the execution finds 98.6\% of the compute is attributed to neighbor search.
In addition, the \texttt{NVIDIA Nsight Compute} tool
shows 46.4\% stream multiprocessor (SM) throughput but only 9.1\% DRAM throughput.
That means the kernel uses about half the compute capacity but barely touching available memory bandwidth,
In addition, although 6.97 warps per scheduler were active on average, only 0.83 were
eligible to issue; no warp was eligible during 55.6\% of cycles.  About 17.1
of 32 lanes were active on average.  The kernel is therefore dominated by
dependent-load latency and divergent, query-dependent work rather than raw
external-memory bandwidth.  This explains why adding grid dimensions or
cooperative search did not help.

In the interleaved public-interface
comparison, the \texttt{frnn\_cuda} NumPy path has a median of 299.6~ms
(95\% CI: 299.1--300.0~ms), whereas the original GPU-resident PyTorch path has
a median of 473.0~ms (95\% CI: 472.3--474.3~ms).  All 30 paired
rounds favor \texttt{frnn\_cuda} (two-sided exact sign test,
$p=1.86\times10^{-9}$), and the ratio of the two medians is 1.58.  Because the
API boundaries differ, this comparison
should be interpreted as an end-to-end interface result rather than a
kernel-only attribution.  Separately, the optimized native-device path reaches
267.8~ms with a reused workspace.  Peak memory on the real workload decreases
from 3.08~GiB to 2.06~GiB.

The same comparison was also performed on NVIDIA A100-SXM4 GPUs at NERSC.
Table~\ref{tab:generalization} summarizes the comparison results.
With the common PyTorch 2.13.0/CUDA 13.0
stack, the 40~GB and 80~GB A100 measurements agree within their confidence
intervals and yield speedups of 1.47 and 1.50, respectively.  The alternative
PyTorch 2.8.0/CUDA 12.9 stack yields a speedup of 1.38.

\begin{table}[htbp]
\centering
\footnotesize
\caption{Real-workload comparison across GPU generations.  Values are
median latencies in milliseconds over 30 interleaved rounds; brackets give
95\% bootstrap confidence intervals.  Speedup is the paired ratio of the
original FRNN latency to the \texttt{frnn\_cuda} latency, with its 95\%
bootstrap interval.  Stack lists the PyTorch and CUDA versions.  Every row
reproduces the complete 9,279,672-edge reference set and favors
\texttt{frnn\_cuda} in all 30 rounds (two-sided exact sign test,
$p=1.86\times10^{-9}$).}
\label{tab:generalization}
\begin{tabular}{llrrr}
\toprule
GPU & Stack & \texttt{frnn\_cuda} host & Original FRNN & Speedup \\
\midrule
RTX 2070 SUPER & 2.13.0 / 13.0 & 299.6 [299.1, 300.0] & 473.0 [472.3, 474.3]
  & 1.58 [1.58, 1.58] \\
A100-SXM4 (40~GB) & 2.13.0 / 13.0 & 166.6 [163.1, 171.1] & 244.7 [244.5, 244.9]
  & 1.47 [1.43, 1.50] \\
A100-SXM4 (80~GB) & 2.13.0 / 13.0 & 164.2 [162.8, 170.8] & 246.7 [246.5, 246.9]
  & 1.50 [1.45, 1.52] \\
A100-SXM4 (40~GB) & 2.8.0 / 12.9 & 177.2 [171.2, 189.1] & 244.7 [244.5, 244.9]
  & 1.38 [1.30, 1.43] \\
\bottomrule
\end{tabular}
\end{table}

\subsection{Rerun agent-directed optimization}
The primary trajectory shows that a goal-persistent agent can carry an
optimization campaign to an exact result, but one run does not show
whether that outcome depended on the particular sequence of hypotheses chosen,
or whether the executable correctness contract stays adequate once the agent
explores different transformations.  To probe both questions, we reran the
continuation stage independently: a fresh Claude Code session started from the
same Codex handoff commit and the same RTX 2070 SUPER hardware, using CUDA
13.0 and native SM75 compilation, but explored its own hypothesis sequence
under the same executable goal.
Table~\ref{tab:claude-r1} summarizes its main experiments before the provider
session limit ended execution after 91.9~min, well below the nominal 5.5~h
budget. It consumed total 10~M input tokens and produced 305~k output tokens.
The fastest candidate reduced warm-device latency from 307.861~ms to
207.299~ms ($p_{95}=208.730$~ms), a 32.7\% reduction, but did not reach the
requested sub-200~ms threshold.
Its baseline latency is different from that in the primary trajectory
even though the source code is identifical. This may be a result
of different software stacks and runtime environments.
Its largest step deferred the indirect
original-index lookup until after radius rejection, reducing the median from
250.198~ms to 207.299~ms.

\begin{table}[htbp]
\centering
\caption{Independent continuation-stage Claude rerun on the real workload.
Values are
warm-device medians.  The final candidate is exact on this workload but its
heap-free dispatch is not exact for arbitrary densities.}
\label{tab:claude-r1}
\begin{tabular}{lrl}
\toprule
Variant & Device $p_{50}$ (ms) & Outcome \\
\midrule
Native SM75 baseline & 307.861 & Reference \\
Warp-cooperative search & 537.598 & Rejected: slower \\
Append then post-sort & 292.526 & Core test passed \\
Grouped D12 distance loads & 252.996 & Core test passed \\
No distance early exit & 268.761 & Rejected: slower \\
Heap-free $K\geq256$ dispatch & 250.198 & Unsafe for dense inputs \\
Deferred original-index lookup & 207.299 & Exact on real workload \\
\bottomrule
\end{tabular}
\end{table}

\section{Limitations}
\label{sec:limitations}

This study presents one primary optimization trajectory and
one continuation-stage rerun rather than a controlled agent evaluation.
Although both Claude runs began from the same Codex handoff,
the primary run accumulated 5~h 30~min of active-turn time,
whereas the rerun ended after 91.9~min because of the provider’s session-usage limit.
The trajectories demonstrate that different optimization paths
can emerge from the same checkpoint,
but their unequal budgets prevent estimation of the expected outcome
across repeated runs or isolation of the effects of goal persistence,
model choice, profiler access, execution environment, and provider constraints.

The rerun also clarifies the scope of the executable correctness contract.
Its differential tests exercised \(K\leq64\), whereas the optimized \(K\geq256\)
path omits heap maintenance.
For the target particle-tracking workload,
every query has fewer than \(K\) in-radius neighbors,
and post-run comparison with the complete reference edge set confirmed exactness.
The optimization is therefore valid for the evaluated particle-tracking workload.
For denser inputs, however, more than \(K\) neighbors may fall within the radius,
in which case preserving general nearest-\(K\) semantics requires heap maintenance.
Extending this workload-specialized optimization into a generally
exact FRNN implementation would require tests covering the new dispatch
and adversarial high-density cases.

The implementation is faster than the original on every tested Turing and
Ampere GPU.  However, the evaluation is limited to single-GPU CUDA
execution and one representative scientific workload.
Evaluating newer GPU generations, broader workload and density regimes,
power efficiency, and multi-GPU scaling would
further characterize its performance portability.

The case study optimizes one algorithmic library, not an end-to-end scientific
workflow.  The mechanism itself uses no FRNN-specific mutation or parameter
operators; the agent reasons over repository state, executable tests, profiler
output, and a persistent completion condition.  These same
interfaces could, in principle, expose workflow-level choices such as data
preparation, algorithm selection, resource placement, and stage boundaries.
Evaluating this extension will require a future study with workflow-level
objectives, correctness constraints, and end-to-end scientific metrics.

\section{Conclusion}

In this case study, general-purpose coding agents sustained an iterative
process of scientific software optimization.
A persistent objective, correctness tests, repository state,
and benchmark feedback guided their experiments.
They tested hypotheses, retained improvements,
rejected regressions, and repaired a correctness issue
found during a later audit.

The resulting standalone C++/CUDA \texttt{libFRNN} implementation reproduces
a representative example of the particle-tracking workload.
Its synchronous NumPy achieved a 1.58-fold speedup
over the original GPU-resident PyTorch path, despite
including host transfers.
The retained design combines several novel ideas, including four-axis gridding,
structure-of-arrays screening, roundoff-safe early rejection, adaptive top-$K$
maintenance, workspace reuse, and compact output.  The experimental record
also contains unsuccessful, rejected hypotheses, including five-dimensional gridding and
warp-cooperative search.

The independent rerun started from the same optimization checkpoint. It explored
a different sequence of hypotheses and found a workload-specialized variant
reaching even better performance.  A subsequent audit showed that this optimization relies on
a density assumption that holds for the target particle-tracking workload but
not for arbitrary FRNN inputs.  Correctness tests and scientific contracts
may evolve with the optimization, since transformations found by
the agent can reveal assumptions that need to be made explicit and tested.

More generally, scientific software optimization can be given to
general-purpose coding agents as an empirical research process.
This study shows that the approach is
feasible for a real GPU algorithm and repository.
Future work can characterize its reproducibility across repeated
runs, broader workloads, and hardware; compare it with alternative optimization
approaches; and extend the same pattern from individual kernels and libraries
to end-to-end scientific workflows.

\section*{Acknowledgements}
This resarch is in support of DOE Award No. DE-SCL0000090 “HEPAmSC IDA Pilot: Knowledge Extraction”.
This research used resources of the National Energy Research Scientific Computing Center
(NERSC), a Department of Energy User Facility using NERSC award HEP-ERCAP 0036579.

\IfFileExists{SciPost_bibstyle.bst}{\bibliographystyle{SciPost_bibstyle}}{\bibliographystyle{unsrt}}
\bibliography{references}

\end{document}